\documentclass{article}
\usepackage{spconf,amsmath,graphicx}
\usepackage{amssymb}
\usepackage{xcolor}
\usepackage{url}
\usepackage{hyperref} 
\usepackage{lipsum}

\newcommand\blfootnote[1]{%
  \begingroup
  \renewcommand\thefootnote{}\footnote{#1}%
  \addtocounter{footnote}{-1}%
  \endgroup
}

\newcounter{ToDo}
\newcounter{gaocomm} 
\newcounter{Note}
\definecolor{blue-violet}{rgb}{0.00,0.75,0.90}
\definecolor{mygreen}{rgb}{0.0, 0.5, 0.0}
\definecolor{awesome}{rgb}{1.0, 0.13, 0.32}
\definecolor{bostonuniversityred}{rgb}{0.8, 0.0, 0.0}

\renewcommand\thefootnote{\textcolor{blue}{\arabic{footnote}}}

\title{Bregman Graph Neural Network}
\name{Jiayu Zhai, Lequan Lin, Dai Shi, and Junbin Gao}
\address{Discipline of Business Analytics, The University of Sydney Business School\\ The University of Sydney, Camperdown, NSW 2006, Australia\\
jzha2776@uni.sydney.edu.au, 
\{lequan.lin, dai.shi, junbin.gao\}@sydney.edu.au}
\begin{document}
%
\maketitle
\begin{abstract}
Numerous recent research on graph neural networks (GNNs) has focused on formulating GNN architectures as an optimization problem with the smoothness assumption. However, in node classification tasks, the smoothing effect induced by GNNs tends to assimilate representations and over-homogenize labels of connected nodes, leading to adverse effects such as over-smoothing and misclassification.
In this paper, we propose a novel bilevel optimization framework for GNNs inspired by the notion of Bregman distance. We demonstrate that the GNN layer proposed accordingly can effectively mitigate the over-smoothing issue by introducing a mechanism reminiscent of the ``skip connection''.
We validate our theoretical results through comprehensive empirical studies in which Bregman-enhanced GNNs outperform their original counterparts in both homophilic and heterophilic graphs. Furthermore, our experiments also show that Bregman GNNs can produce more robust learning accuracy even when the number of layers is high, suggesting the effectiveness of the proposed method in alleviating the over-smoothing issue.
\blfootnote{\copyright 2023 IEEE. Personal use of this material is permitted. Permission from IEEE must be obtained for all other uses, in any current or future media, including reprinting/republishing this material for advertising or promotional purposes, creating new collective works, for resale or redistribution to servers or lists, or reuse of any copyrighted component of this work in other works.}

\end{abstract}

\section{Introduction}\label{Sec:1}
\label{sec:intro}
With the extraordinary ability to encode complex relationships among entities in a system, graph data are widely observed in many application domains, such as social networks \cite{fan2019graph, social1}, biological networks \cite{bio1}, recommender systems \cite{wu2022graph,ying2018graph}, transportation networks \cite{zheng2020gman, trans1}, etc. Graphs normally model entities as nodes and then construct edges between node pairs to represent underlying relationships. In addition, node attributes are represented as graph signals. Traditional deep feed-forward neural networks (NNs) only consider the propagation of features (i.e., columns of the graph signal matrix), which leaves the connectivity among nodes unexploited. To overcome this limitation, graph neural networks (GNNs) are designed to additionally aggregate neighbouring node features in the direction of rows, contributing to better graph representation learning (GRL) and eventually outstanding predictive performance in various tasks \cite{zhou2020graph}.

Framing NNs as an optimization problem is a well-established research topic in the machine learning community \cite{frecon2022bregman, yang2021robust, louati2021deep}. Likewise, numerous recent research on GNNs focuses on the optimization formulation of GNN layers or the end-to-end GNN training. Some works have shown that GRL can be approximated by the solution of some optimization problem with the smoothness assumption of neighbouring node representations \cite{zhu2021interpreting, ma2021unified, yang2021attributes}. It has also been proven that the end-to-end training for GNNs can be formulated as a bilevel optimization problem, or alternatively, a faster multi-view single-level optimization framework \cite{han2023alternately}. In this work, we will consider the bilevel optimization formulation of GNNs, in which the upper-level problem shares the same purpose as optimizing the objective function, and the lower-level problem conducts GRL.

Unifying GNNs as optimization problems provides a new perspective to understanding and analyzing existing methods. For example, considering GNNs in node classification tasks, the smoothness assumption, which tends to homogenize the labels of connected nodes, can lead to several adverse effects such as over-smoothing and inappropriate message-passing for heterophilic graphs \cite{yang2021attributes, yan2022two}. 
Specifically, the so-called over-smoothing issue appears when node features become indistinguishable after several propagations of GNN layers. This phenomenon is more evident in the graphs where connected nodes are often with the same label, known as homophily. 
On the other hand, with heterophilic graphs where connected nodes have different labels, the smoothing effect induced by GNNs can even lead to worse classification outcomes, because the model is prone to assign similar labels to connect nodes with similar features after smoothing.

The above-mentioned issues can be mitigated with the concept of ``skip connection'' \cite{he2016resnethu}. For example, APPNP \cite{KlicperaBG19} combines the original node feature with the representation learned by each layer, which effectively preserves local information and helps mitigate over-smoothing issues. Such methods are also helpful with heterophilic graphs because they mitigate the effect of smoothing in representation learning. It has been shown that designing NNs as a bilevel optimization problem with penalty on the Bregman distance between representations from each two consecutive layers is reminiscent of and even better than applying skip connection \cite{frecon2022bregman, dhillon2008matrix}. This method simplifies the network architecture by employing a set of invertible activation functions. However, it has no direct extension to GNNs as the problem design is limited by the feature propagation of traditional NNs.

In this paper, we aim to propose a novel bilevel optimization framework for GNNs enlightened by the notion of Bregman distance that can effectively alleviate the adverse effects of smoothing. Similar to other bilevel designs, we develop the upper-level problem to optimize the overall objective function, and the lower-level problem for GRL. We show that the optimization framework can be easily applied to the computational format of GNNs by introducing the same set of activation functions for Bregman NNs \cite{frecon2022bregman}, and we name such architectures as Bregman GNNs.

The contributions of this work include (1) a novel bilevel optimization framework for designing GNNs with Bregman distance; (2) an alternative solution to the adverse effects of smoothing with a set of specially-designed activation functions sharing a similar purpose with skip connection;  (3) solid numerical experiment results to validate the effectiveness of the new framework.

\section{The Proposed Framework}\label{Sec:2}


\subsection{Preliminaries}\label{SubSec:2.1}
We denote  $\mathcal{G}=(\mathcal{V},\mathcal{E})$ as an undirected graph with a set of nodes $\mathcal{V}$ and a set of edges $\mathcal{E}$. $\mathbf Z^l \in \mathbb{R}^{n \times d_l}$ denotes the node feature matrix at layer $l$, where $n$ is the number of nodes, and $d_l$ is the embedding size. The graph adjacency matrix is denoted as $\mathbf A \in \mathbb{R}^{n \times n}$ with 
$\mathbf A_{ij}=1$ if node $i$ is connected with $j$, and $\mathbf A_{ij}=0$ otherwise.
We further let $\mathbf D\in \mathbb{R}^{n \times n}$ be the degree matrix, where $d_i = \sum_j \mathbf A_{ij}$. We now provide some necessary notations and definitions for the formulation of Bregman GNN layers.

\noindent \textbf{Definition 1} (Class of layer-wise functions $\mathcal{F}$ \cite{frecon2022bregman}). 
 Define $\{f_l\}^{L-1}_{l=0}: \mathbb{R}^{d_{l+1}}\times \mathbb{R}^{d_{l+1}} \rightarrow \mathbb{R}$ to be a specific set of bi-linear functions such that
\begin{align}\label{initial_defn_f}
    f_l(\mathbf z,\mathbf z_i^l \mathbf M_l) = (\mathbf z_i^{l}\mathbf M_l)^\top \mathbf E_l^\top \mathbf z - \! \mathbf b_l^\top\mathbf z - \! \mathbf c_l^\top(\mathbf z_i^l\mathbf M_l) + {\delta}_l ,
\end{align}
where $\mathbf b_l, \mathbf c_l \in \mathbb{R}^{d_{l+1}}$ and ${\delta}_l\in \mathbb{R}$. $\mathbf{z}_i \in \mathbb{R}^{d_{l}}$ and $\mathbf{z} \in \mathbb{R}^{d_{l+1}}$ are the feature vectors of sample $i$ at layer $l$ and $l+1$, respectively. Finally, the matrix $\mathbf M_l \in \mathbb R^{d_l\times d_{l+1}}$ is the weight matrix, and $\mathbf{E}_l \in \mathbb{R}^{d_{l+1}\times d_{l+1}} $ is the parameter matrix presenting the feature correlation. We note that such design of $\mathcal F$ guarantees the closed form solution of the lower-level optimization of the problem defined in Eq.~\eqref{nnopt}, and this form of $\mathcal F$ has been further assigned to enhance the performance of NN in the work of \cite{frecon2022bregman}. We now show how to establish the notion of $\mathcal F$ to the \textbf{graph data}. Rather than the feature propagation in NN, where features are considered individually as single vectors, in GNN one shall require to propagate the feature matrix as a whole due to the connectivity of the nodes. Accordingly, one shall consider assigning the matrix trace for each of the terms of the definition of $\mathcal F$, resulting in the following form: 
\begin{align}\label{f_form_in_graph}
f_l(\mathbf Z, \mathbf A \mathbf Z^l\mathbf M_l) &= \text{tr}((\mathbf A \mathbf Z^l \mathbf M_l)\mathbf E_l \mathbf Z^\top) - \langle \mathbf{1} \times \mathbf b_l^\top,\mathbf Z \rangle \notag \\ &- \langle \mathbf{1} \times \mathbf c_l^\top, \mathbf A \mathbf Z^l \mathbf M_l \rangle + {\delta}_l ,
\end{align}
where $\mathbf{1}$ is the $n$-dimensional vector with all ones, and $\langle \cdot,\cdot \rangle$ is the inner product between two matrices. We note that the inclusion of the inner product is due to the fact $\langle \mathbf A, \mathbf B\rangle = \mathrm{tr}(\mathbf A^\top \mathbf B)$.  Similarly, as we will show in Section \ref{SubSec:2.2}, the form of $\mathcal F$ under Eq.~\eqref{f_form_in_graph} also guarantees closed form solution of the low-level optimization problem defined in Eq.~\eqref{gnnopt} for GNN. Additionally, to properly define the Bregman GNN layer, we further provide the notion of Bregman distance and proximity operator as follows.
\\[7pt]
\textbf{Definition 2} (Bregman distance\cite{dhillon2008matrix}).
Bregman distance of the matrix $\mathbf{P}$ from the matrix $\mathbf{Q}$ is
$$D_{\phi} (\mathbf P, \mathbf Q)=\phi(\mathbf P)-\phi(\mathbf Q)- \langle \nabla \phi (\mathbf Q), \mathbf P-\mathbf Q \rangle,$$
where $\phi$ is a Legendre function \cite{bauschke2018breg}. The Bregman distance is actually a general case of many distance measurements. For example, if $\phi(\mathbf{P})=\frac{1}{2} \lVert \mathbf P \rVert^2$, then $D_{\phi} (\mathbf P, \mathbf Q)=\frac{1}{2} \lVert \mathbf P-\mathbf Q \rVert^2$ is the square Euclidean distance.
\\[7pt]
\textbf{Definition 3} (Bregman proximity operator \cite{van2017forward}).
The Bregman proximity operator of $g$ with respect to $\phi$ is denoted by
\begin{align}\label{proximity_operator}
    \operatorname{prox}_g^{\phi}(\mathbf P)=\underset{\mathbf Q}{\operatorname{argmin}} \{g(\mathbf Q)+\phi(\mathbf Q)-\langle \mathbf Q, \mathbf P\rangle \}.
\end{align}
In the next section, we show how bilevel optimization can be constructed for graph data based on these definitions.

\subsection{Bilevel optimization for graph data}\label{SubSec:2.2}

We start by recalling bilevel optimization on the data input (i.e., images) in general NN. Given a standard training data set $\{\mathbf x_i, \mathbf y_i\}_{i=1}^n$ where $\{\mathbf x_i, \mathbf y_i\} \in \mathbb{R}^{d_0}\times \mathbb{R}^{c}$, one
can denote the feature propagation of NN as the bilevel optimization problem as follows \cite{frecon2022bregman}.
\begin{equation} \label{nnopt}
\begin{gathered}
\underset{\substack{\psi, \left\{f_l\right\}_{l=0}^{L-1}}}
{\operatorname{minimize}} \quad \sum_{i=1}^n \ell\left(\psi\left(\mathbf z_i^{L}\right), \mathbf y_i\right) 
\quad \text { where } \forall i \in[n],\\
\quad\left\{\begin{array}{l}
\mathbf z_i^{0}=\mathbf x_i \\
\text { for } l=0,1, \ldots, L-1, \\
\mathbf z_i^{l+1}=\underset{\mathbf z \in \mathbb{R}^{d_{l+1}}}{\operatorname{argmin}} \{f_l\left(\mathbf z, \mathbf z_i^{l}\mathbf M_l\right) +D\left(\mathbf z, \mathbf z_i^{l}\mathbf M_l\right)+g(\mathbf z)\},
\end{array}\right.
\end{gathered} 
\end{equation}
where $\psi \in \mathcal{B}\left(\mathbb{R}^{d_{L}}, \mathbb{R}^c\right)$ serving as Borel measurable function, $\left\{f_l\right\}_{l=0}^{L-1} \in \mathcal{F}\left(\mathbb{R}^{d_{l+1}} \times \mathbb{R}^{d_{l+1}}\right)^L$, and $g\in \Gamma_0(\mathbb R^d)$ can be treated as simple convex function for regularization. 
The upper-level objective is the loss between the prediction $\widehat{\mathbf {y}}_i=\psi(\mathbf z_i^{L})$ and the ground truth of $\mathbf y_i$, where $\psi $ is a simple transformation such as a linear layer or a linear layer followed by a softmax operator. $\ell$ is the loss function, such as cross-entropy for classification tasks and quadratic loss for regression. The lower-level optimization problem produces layer-wise feature representation and can be further unrolled as an NN layer \cite{frecon2022bregman}.

Now we analogize the notion of bilevel optimization from the scope of NN to the graph structured data. It is well-known that the core difference between the propagation in NN and GNN is whether the connectivity between nodes (or samples) is considered \cite{kipf2017semisupervised}. Specifically, unlike NN in which each node feature is propagated individually, the neighbouring information is gathered for each node via GNN propagation according to graph connectivity (i.e., adjacency matrix $\mathbf A$). Therefore, it is natural for one to generalize the bilevel optimization process defined in Eq.~\eqref{nnopt} by including the graph adjacency information. Accordingly, the lower-level objective becomes
\begin{align}  \label{gnnopt}
\mathbf Z^{l+1} \!\!= \!\!\!\! \underset{\mathbf Z \in \mathbb{R}^{n \times d_{l+1}}}{\operatorname{argmin}} \!\!\!  \{\! f_l(\mathbf Z, \mathbf A \mathbf Z^l \mathbf M_l)  + D_{\phi_l} (\mathbf Z, \mathbf A \mathbf Z^l \mathbf M_l) \! +\! g_l(\mathbf Z)\!\}.
\end{align}
It is not difficult to verify that with the form of $f$ defined in \textbf{Definition 1}, the optimization above can still have a closed form solution. 
The second term measures the closeness between the feature vectors in layer $l$ and $l+1$. The minimization of such term restricts the changes in the feature matrix between layers, thereby diluting the smoothing effects. 
\\[7pt]
\textbf{Remark 1}.
The form of the $f_l$ can be seen as the negative energy in Restricted Boltzmann Machine \cite{hinton2012practical}. The energy between two vectors $\mathbf{u}$ and  $\mathbf{v}$ is defined as:
$$E(\mathbf{u}, \mathbf{v})=-\mathbf{u}^\top \mathbf{E} \mathbf{v} - \mathbf{b}^\top \mathbf{u}-\mathbf{c}^\top\mathbf{v}.$$
Thus, the optimization problem aims to maximize this energy.

\begin{figure}[t]
\begin{minipage}[b]{1.0\linewidth}
  \centering
\centerline{\includegraphics[width=9cm]{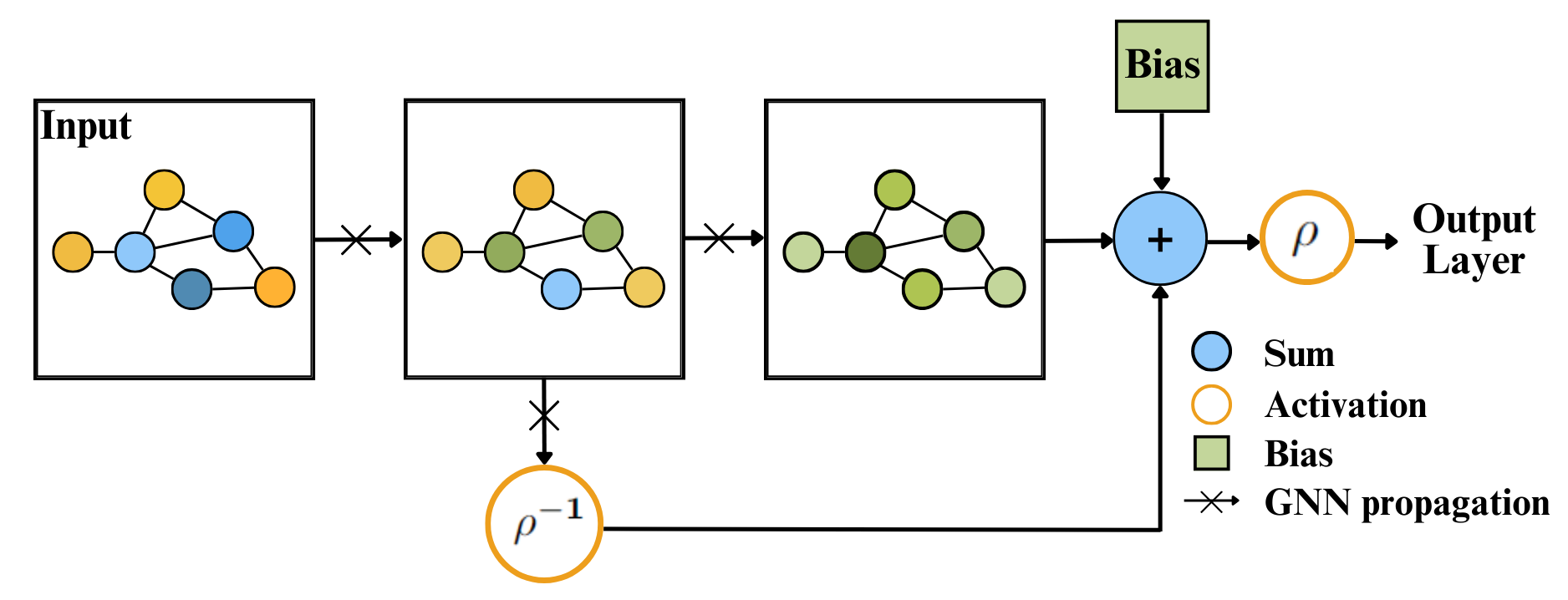}}
\end{minipage}
\caption{Illustration of Bregman GNN based on Equation (\ref{eq4}). The model is composed of one hidden layer of classic GNN propagation, one hidden layer of Bregman-modified propagation with the invertible activation functions, and finally, the output layer to make predictions. This architecture can be extended by adding more hidden layers.}
\label{fig:method}
\end{figure}

\begin{table}
\centering
\footnotesize
\caption{Statistics of the homophilic and heterophilic datasets}
    \label{Data_stat}
    \setlength{\tabcolsep}{1mm}{
    \begin{tabular}{ccccccc}
    \hline
        \textbf{Datasets} & Class & Feature & Node & Edge & Train/val/test & Homophily\%\\
        \hline
        \textbf{Cora} & 7 & 1433 & 2708 & 5278 & 140/500/1000 & 82.5\%\\
        \textbf{CiteSeer} & 6 & 3703 & 3327 & 4552 & 120/500/1000 & 72.1\%\\
        \hline
        \textbf{Actor} &5 &932 &7600 &26659 & 60\%/20\%/20\% & 21.4\%\\
        \textbf{Texas} &5 &1703 &183 &279 &60\%/20\%/20\% & 11.0\%\\
        \hline
    \end{tabular}}
\end{table}

\begin{table*}
\footnotesize
\caption{Comparison Experiment Results: Node Classification Accuracy (\%)}
    \label{Table_results}
    \centering
    \setlength{\tabcolsep}{1mm}{
    \begin{tabular}{c|c|c|c|c|c|c|c|c}\hline
          & \multicolumn{2}{c}{\textbf{Cora}} \vline & \multicolumn{2}{c}{\textbf{CiteSeer}} \vline & \multicolumn{2}{c}{\textbf{Texas}}\vline & \multicolumn{2}{c}{\textbf{Actor}}\\
         \hline
          \textbf{Models} & \textbf{Bregman} & \textbf{Standard} &  \textbf{Bregman} & \textbf{Standard} & \textbf{Bregman} & \textbf{Standard} & \textbf{Bregman} & \textbf{Standard}  \\
        \hline
        ChebNet \cite{defferrard2016convolutional}
        & 81.22 $\pm$ 0.94 & \textbf{81.46 $\pm$ 0.54} & \textbf{71.70 $\pm$ 0.50} & 71.68 $\pm$ 1.20 & \textbf{84.05 $\pm$ 5.47} & 83.51 $\pm$ 3.91 & \textbf{35.92 $\pm$ 0.84} & 35.81 $\pm$ 1.16 \\
        
        GCN \cite{kipf2017semisupervised}
        & \textbf{82.58 $\pm$ 0.84} & 82.32 $\pm$ 0.69 & \textbf{72.35 $\pm$ 0.85} & 71.51 $\pm$ 0.40 & \textbf{63.78 $\pm$ 5.31} & 63.24 $\pm$ 4.55 & \textbf{29.05 $\pm$ 0.68} & 27.93 $\pm$ 0.79 \\
        
        GAT \cite{velivckovic2018graph}
        & \textbf{82.19 $\pm$ 0.69} & 81.63 $\pm$ 0.71 & \textbf{71.52 $\pm$ 0.72} & 70.31 $\pm$ 0.81 & \textbf{63.24 $\pm$ 3.86} & 62.70 $\pm$ 3.15 & \textbf{29.45 $\pm$ 0.52} & 28.48 $\pm$ 0.70 \\
        
        APPNP \cite{KlicperaBG19}
        & \textbf{82.27 $\pm$ 0.63} & 80.70 $\pm$ 0.66 & \textbf{72.67 $\pm$ 0.60} & 71.30 $\pm$ 0.78 & 61.62 $\pm$ 4.65 & \textbf{62.70 $\pm$ 5.10} & \textbf{27.27 $\pm$ 0.97} & 26.19 $\pm$ 1.16 \\
        
        GIN \cite{xu2018how}
        & \textbf{80.36 $\pm$ 0.76} & 80.04 $\pm$ 1.26 & \textbf{69.82 $\pm$ 0.79} & 69.23 $\pm$ 0.79 & \textbf{63.51 $\pm$ 5.02} & 63.24 $\pm$ 5.30 & \textbf{28.47 $\pm$ 1.04} & 27.43 $\pm$ 1.19 \\

        GraphSAGE \cite{hamilton2017inductive}
        & \textbf{81.74 $\pm$ 0.66} & 81.63 $\pm$ 0.47 & \textbf{70.81 $\pm$ 0.57} & 70.53 $\pm$ 0.85 & 83.51 $\pm$ 4.75 & \textbf{83.78 $\pm$ 3.82} & 35.34 $\pm$ 0.68 & \textbf{35.60 $\pm$ 0.70} \\
        \hline
    \end{tabular}}
\end{table*}

\subsection{Bregman GNN layers}


In this section, we show how the Bregman GNN layer is built. A demonstration of model architecture is provided in \textbf{Fig. \ref{fig:method}}. According to Frecon et al. \cite{frecon2022bregman}, many widely used activation functions (e.g., Relu and Arctan) can be written as the inverse gradient of strongly convex Legendre functions $\phi$, and for some particular choice of $g$ and $\phi$, the Bregman proximity operator in Eq.~\eqref{proximity_operator} can be written as
$$\operatorname{prox}_g^{\phi}(\mathbf{P}) = \nabla \phi^{-1}(\mathbf{P})=\rho (\mathbf{P}) .$$
Since $\mathrm{tr}((\mathbf A \mathbf Z^l \mathbf M_l)\mathbf E_l \mathbf Z^\top)=\langle (\mathbf A \mathbf Z^l \mathbf M_l)\mathbf E_l, \mathbf Z\rangle$, Eq.~\eqref{gnnopt} becomes
\begin{equation} \label{deri}
\begin{split}
\mathbf Z^{l+1} \!\! & = \!\!\! \underset{\mathbf Z \in \mathbb{R}^{n \times d_{l+1}}}{\operatorname{argmin}}  \{ f_l(\mathbf Z, \mathbf A \mathbf Z^l \mathbf M_l) \! + \! D_{\phi_l} (\mathbf Z, \mathbf A \mathbf Z^l \mathbf M_l)\!+ \! g_l(\mathbf Z)\}\\
 & = \underset{\mathbf Z \in \mathbb{R}^{n \times d_{l+1}}}{\operatorname{argmin}} \{g_l(\mathbf Z) +\phi(\mathbf Z) - \langle \nabla \phi(\mathbf A \mathbf Z^l \mathbf M_l)\\ 
 &\quad -(\mathbf A \mathbf Z^l\mathbf M_l) \mathbf E_l+\mathbf{1} \times \mathbf b_l^\top, \mathbf Z\rangle \}\\
 & = \text{prox}_g^\phi (\nabla \phi(\mathbf A \mathbf Z^l \mathbf M_l)-(\mathbf A \mathbf Z^l \mathbf W_l) \mathbf E_l+\mathbf{1} \times \mathbf b_l^\top)\\
 & = \rho(\rho^{-1}(\mathbf A \mathbf Z^l \mathbf M_l)-(\mathbf A \mathbf Z^l \mathbf W_l) \mathbf E_l+\mathbf{1} \times \mathbf b_l^\top), 
\end{split}
\end{equation}
where we have $D_{\phi_l} (\mathbf Z, \mathbf A\mathbf Z^l \mathbf M_l)
 =\phi_l(\mathbf Z)-\phi_l(\mathbf A \mathbf Z^l \mathbf M_l)- \langle \nabla \phi_l (\mathbf A \mathbf Z^l \mathbf M_l), \mathbf Z-\mathbf A \mathbf Z^l \mathbf M_l \rangle$. If we furhter let $\mathbf W_l=-\mathbf E_l \in \mathbb{R}^{d_{l+1} \times d_{l+1}}$ be the weight matrix, and $\mathbf b_l \in \mathbb{R}^{d_{l+1}}$be the bias, then Eq.~\eqref{deri} can be seen as a layer of GNN:  
\begin{equation} \label{eq4}
\begin{split}
\mathbf Z^{l+1} = \rho(\rho^{-1}(\mathbf A \mathbf Z^l \mathbf M_l)+\mathbf A \mathbf Z^l \mathbf W_l+\mathbf{1}\times \mathbf b_l^\top),
\end{split}
\end{equation}
where $\rho$ is the activation function and $\rho^{-1}$ is its inverse.
If $\mathbf Z^l$ and $\mathbf Z^{l+1}$ share the same dimension i.e., $n \times d_l$, then $\mathbf M_l \in \mathbb R^{d_l \times d_l}$. 
$\mathbf W_l \in \mathbb{R}^{d_{l}\times d_{l+1}}$ represents the weights in layer $l$.
$\mathbf b_l \in \mathbb{R}^{d_{l+1}\times d_{l+1}}$ represents the biases in layer $l$.
Hence, the parameters that the model should learn are $\mathbf M_l$, $\mathbf W_l$, and $\mathbf b_l$.
Regarding the term $\rho^{-1}(\mathbf A \mathbf Z^l \mathbf M_l)$ in the derivation of Eq.~\eqref{deri}, the utilization of inverse activation function for $\mathbf A \mathbf Z^l \mathbf M_l$ brings the feature representation of the previous layer to the present layer. This serves a similar purpose as skip connection. Therefore, such design helps the model to maintain the desirable variation of node features, thus inducing the adverse effect of smoothing in GNN propagation. Finally, it is worth noting that the propagation in Eq.~\eqref{eq4} can be applied to many existing spatial message-passing GNNs such as GCN \cite{kipf2017semisupervised} and GAT \cite{velivckovic2018graph}. In the next section, we verify such enhancement power of Eq.~\eqref{eq4} with comprehensive empirical studies. 

\section{Experiments}\label{Sec:3}
The primary objective of our experiment is to test the performance of the proposed Bregman GNNs in comparison with their standard forms, which means the experiments are conducted in an ablation fashion. 
We first compare the performance of Bregman-enhanced GNNs to their standard forms to show their adaptive power on both homophily and heterophily graphs.
Then, we provide the results of an over-smoothing experiment to show the effectiveness of the proposed method in alleviating over-smoothing. Our experiment codes can be found at \url{https://github.com/jiayuzhai1207/BregmanGNN}.

\subsection{Datasets and Implementation Details}
For the first experiment, we choose 4 commonly-used datasets as shown in \textbf{Table \ref{Data_stat}}, including 2 homophilic graphs \textbf{Cora} \cite{yang2016revisiting} and \textbf{CiteSeer} \cite{yang2016revisiting}, and 2 heterophilic graphs \textbf{Actor} \cite{Pei2020Geom-GCN} and \textbf{Texas} \cite{Pei2020Geom-GCN}. For the over-smoothing experiment, we only use \textbf{Actor}. 
The train/validation/test split follows the same split in  \cite{yang2016revisiting} and \cite{Pei2020Geom-GCN}.
In the first experiment, we choose 6 classic GNNs as baselines, and all networks have 3 layers including the output layer. We select this architecture because Bregman GNNs require at least 3 layers: 2 hidden layers to apply the inverse activation function, and the output layer for final classification. In the over-smoothing experiment, we choose GCN and GAT as baselines and set the number of layers in $[3, 5, 7, 9]$.
The average test accuracy and its standard deviation are calculated based on the results from 10 runs. Grid search is conducted for hyperparameter tuning.
For Bregman GNNs, we select from a set of invertible activation functions that have been shown as Bregman proximity operators, such as ReLU, Tanh, ArcTan, and Softplus
\cite{frecon2022bregman}.

\subsection{Results for Homophilic and Heterophilic Graphs}
The experiment results are shown in \textbf{Table \ref{Table_results}}. Overall, Bregman GNNs present good performance compared to their standard counterpart across all datasets. For homophilic graphs, the Bregman architecture achieves consistent improvement on the standard baselines. Notably, the Bregman architecture enhances the accuracy of APPNP by 1.57\% for \textbf{Cora} and by 1.37\% for \textbf{Citeseer}. For heterophilic graphs, the Bregman architecture successfully improves the performance of ChebNet, GCN, and GAT for both \textbf{Texas} and \textbf{Actor} by 0.19\% to 1.12\%. APPNP remains to have the largest improvement from the Bregman enhancement for \textbf{Actor}. One possible reason is that the Bregman architecture provides one additional path for APPNP propagation to access source terms from previous layers, which further mitigates the adverse effect of smoothing. However, no improvement is observed between GraphSAGE and its Bregman form, yet the learning accuracy remains comparable between them. Finally, in most cases, Bregman GNNs show lower standard deviation, indicating higher stability in the node classification task.

\begin{figure}[t]
\begin{minipage}[b]{1.0\linewidth}
  \centering
\centerline{\includegraphics[width=8.5cm, height = 3.7cm]{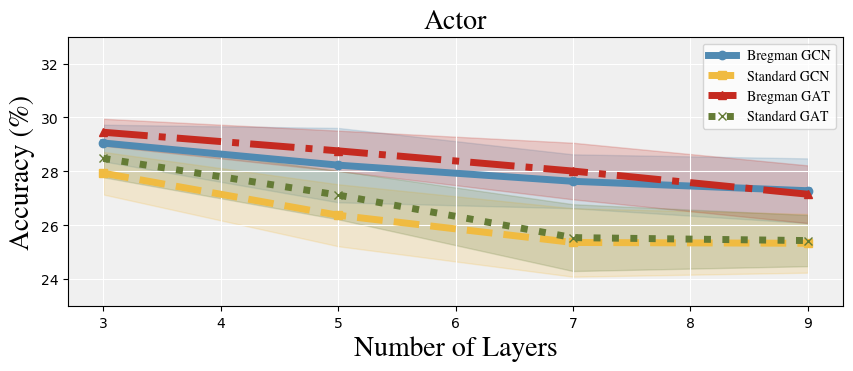}}
\end{minipage}
\caption{Results on Actor for GCN and GAT with different number of layers. Bregman-enhanced GNNs show higher accuracy when the number of layers increases.}
\label{fig:oversmotthing}
\end{figure}

\subsection{Results for Over-smoothing Experiment}
The experiment results are presented in \textbf{Fig. \ref{fig:oversmotthing}}. Apparently, the classification accuracy of both Bregman GCN and GAT is consistently higher than their standard counterparts when the number of layers increases. Therefore, we conclude that Bregman GNNs are more robust to the over-smoothing issue. Nevertheless, the overall decrease trend in accuracy indicates that the over-smoothing issue is only alleviated but not fully addressed.

\section{Conclusion}
\label{sec:conclusion}
In this paper, we have proposed a novel bilevel optimization framework whose closed-form solution naturally defines a set of new network architectures called Bregman GNNs. Our experiments show the proposed framework can improve the performance of classic GNNs on both homophilic and heterophilic graphs and alleviate over-smoothing. However, it is worth noting that our method can only serve as a moderator and cannot fully address the over-smoothing issue. Future works may consider further improvement on this limitation.


\bibliographystyle{IEEEbib}
\bibliography{refs}

\end{document}